\documentclass[10pt,twocolumn,letterpaper]{article}

\usepackage{iccv}
\usepackage{times}
\usepackage{epsfig}
\usepackage{graphicx}
\usepackage{amsmath}
\usepackage{amssymb}
\usepackage{booktabs}
\usepackage{multirow}

\usepackage{algorithm}
\usepackage{algpseudocode}

\usepackage{color}
\definecolor{citecolor}{RGB}{119,215,0} 
\usepackage[pagebackref=false,breaklinks=true,letterpaper=true,colorlinks,citecolor=citecolor,bookmarks=false]{hyperref}%
\iccvfinalcopy 

\def\iccvPaperID{****} 

\begin{document}

\title{Understanding Image Retrieval Re-Ranking: \\ A Graph Neural Network Perspective}


\author{
 Xuanmeng Zhang${^{1,2}}$ \quad  Minyue Jiang${^{2}}$ \quad  Zhedong Zheng${^{1}}$  \\
 Xiao Tan${^{2}}$\quad Errui Ding${^{2}}$ \quad Yi Yang${^{1}}$ \quad \\ 
 $^1$ReLER, University of Technology Sydney, Australia\\  
 $^2$Baidu Inc., China  \\
   
    
}

\maketitle

\begin{abstract}
The re-ranking approach leverages high-confidence retrieved samples to refine retrieval results, which have been widely adopted as a post-processing tool for image retrieval tasks. However, we notice one main flaw of re-ranking, \ie, high computational complexity, which leads to an unaffordable time cost for real-world applications. In this paper, we revisit re-ranking and demonstrate that re-ranking can be reformulated as a high-parallelism Graph Neural Network (GNN) function. 
In particular, we divide the conventional re-ranking process into two phases, \ie, retrieving high-quality gallery samples and updating features. We argue that the first phase equals building the $k$-nearest neighbor graph, while the second phase can be viewed as spreading the message within the graph. In practice, GNN only needs to concern vertices with the connected edges. Since the graph is sparse, we can efficiently update the vertex features. On the Market-1501 dataset, we accelerate the re-ranking processing from \textbf{89.2s} to \textbf{9.4ms} with one K40m GPU, facilitating the real-time post-processing. Similarly, we observe that our method achieves comparable or even better retrieval results on the other four image retrieval benchmarks, \ie, VeRi-776, Oxford-5k, Paris-6k and University-1652, with limited time cost. Our code is publicly available.~\footnote{ \tiny \url{https://github.com/Xuanmeng-Zhang/gnn-re-ranking}}
\end{abstract}

\section{Introduction}
Re-ranking leverages high-confidence retrieved samples to rank the initial retrieval result again~\cite{chum2007total, jegou2007contextual, qin2011hello}, which is usually viewed as a post-processing tool.
It has been widely adopted in various of image retrieval tasks, such as person re-identification~\cite{zhong2017re, saquib2018pose,zheng2018pedestrian},
vehicle re-identification~\cite{huang2019multi,zheng2020vehiclenet} and localization~\cite{shen2012object,zheng2020university}. 
Re-ranking methods can be divided into two categories according to similarity criteria, \ie, feature similarity~\cite{chum2007total, radenovic2018fine} and neighbor similarity~\cite{bai2016sparse, zhong2017re}. 
Given a pair of images, feature similarity is evaluated based on the Euclidean distance in the feature space. In contrast, neighbor similarity measures the number of common neighbors.
For instance, if two samples share more neighbors, they will obtain a higher neighbor similarity value. 
Generally, the neighbor-based methods outperform the feature-based methods. It is because the neighbor-based method is robust to the hard negative (``outliers''), which usually shares one different neighbor set with the true-matches. Considering the robustness to false-matches, we mainly study the neighbor-based re-ranking methods. However, one challenging problem remains. Despite the effectiveness of neighbor-based algorithms, high computation complexity makes this line of approaches unaffordable for real-world applications. 
For instance, the $k$-reciprocal re-ranking method~\cite{zhong2017re} adopts a reciprocal-neighbor rule, which demands the consideration of the neighbors' neighbor and introduces extra computations.

To address the problem of unaffordable high complexity, we consider the feasibility of the parallel inference. In this paper, inspired by the effectiveness and efficiency of graph neural network~(GNN) with the structured data, 
we argue that the re-ranking methods can be reformulated as a high-parallelism GNN function~\cite{gori2005new, bruna2013spectral} to efficiently conduct the re-ranking operation, \eg, neighbor calculation and query expansion. Graph neural network (GNN) draws wide attention in the last few years~\cite{scarselli2008graph}. The increasing interest is due to two main factors: the dominance of data with topology structures in real-world applications, and the limited performance of convolutional neural network (CNN) when dealing with such data. 
Different from the convolution operation of CNN that computation can only take place within local regions, the GNN propagates the features of nodes and edges over the entire graph. 

In particular, the re-ranking process can be divided into two phases, \ie, retrieving high-confidence gallery samples and updating features. 
In the first phase, the conventional re-ranking is to find the high-confidence samples according to the similarity score. For the second phase, the high-confidence samples, \eg, $k$-nearest neighbors, are leveraged to conduct the query expansion by aggregating the feature of these samples. 
Following the spirit of conventional re-ranking, our work first builds the $k$-nearest neighbors ($k$-NN) graph, capturing the topology structure among data. Secondly, we employ the GNN to propagate the message among the whole graph. GNN only needs to update vertices with the connected edges. Due to the sparseness of the $k$-NN graph, we can efficiently update the vertex features. 
On the Market-1501 dataset~\cite{zheng2015scalable}, we accelerate the re-ranking processing from \textbf{89.2s} to \textbf{9.4ms} on GPU, facilitating the real-time post-processing for image retrieval tasks. Furthermore, we observe similar acceleration results on other benchmarks while maintaining competitive performance.
Overall, the main contributions of this work are summarized
as follows:
\begin{enumerate}
\item  
We identify the challenging problem, \ie, large time cost due to high computation complexity, in applying the re-ranking approaches to the real-world scenarios. To address this limitation, we revisit re-ranking methods and demonstrate that the re-ranking process can be re-formulated as high-parallelism Graph Neural Network (GNN), which facilitates the real-time post-processing for image retrieval tasks. 

\item 
Extensive experiments on five datasets, \ie, Market-1501~\cite{zheng2015scalable}, VeRi-776~\cite{liu2016deep}, Oxford-5k~\cite{philbin2007object}, Paris-6k~\cite{philbin2008lost} and University-1652~\cite{zheng2020university}, show the proposed method can significantly accelerate the re-ranking process, while achieving competitive re-ranking results with other conventional re-ranking methods.
\end{enumerate}

\section{Related Works}

\subsection{Re-ranking for Image Retrieval} 
The re-ranking method is one of the common post-processing methods for image retrieval, which can be divided into two categories according to similarity criteria, \ie, feature similarity and neighbor similarity. By taking the nearest samples with similar features into consideration, feature similarity-based methods enrich the query feature by aggregating the features of neighbors. The average query expansion(AQE)~\cite{chum2007total} 
directly averages the top-$k$ similar gallery features as the new query feature. Radenovic~\etal~\cite{radenovic2018fine} propose $\alpha$-weighted query expansion, which assigns different weights to the gallery samples. Lin~\etal~\cite{lin2020bayesian} propose a Bayesian model to predict true matches in the gallery. 
In contrast, several researchers resort to mine the robust neighbor information to improve the accuracy of image retrieval 
\cite{jegou2008accurate, bai2009learning, kontschieder2009beyond, wang2014one}.
Sparse contextual activation (SCA)~\cite{bai2016sparse} is proposed to encode the local contextual distribution of images under the generalized Jaccard metric.~\cite{zhong2017re} design an effective $k$-reciprocal re-ranking method, which inherits the advantages of $k$-reciprocal nearest neighbors~\cite{jegou2007contextual, qin2011hello} and SCA~\cite{bai2016sparse}.
Besides, another line of approaches needs extra annotations. 
Some of these algorithms require human interaction ~\cite{yamamoto2007rerank, rohini2007novel, leng2015person,liu2013pop} or label supervision~\cite{bai2017scalable}.
Furthermore, several methods are based on the ensemble of different ranking metrics to refine the final list. 
For instance, multiple distance measure fusion~\cite{zhang2012automatic, zhang2014query}, rank aggregation~\cite{pedronette2013image, pedronette2014unsupervised} and ranking list comparison~\cite{webber2010similarity} also have achieved great success in the retrieval tasks. 
In general, most re-ranking algorithms pay attention to the ranking performance but neglect the computation efficiency for real-world applications. 
Different from existing methods, we mainly study the efficiency of re-ranking approaches and intend to simultaneously achieve competitive performance and high computation efficiency.

\subsection{Graph Neural Networks} 
Graph Neural Networks~(GNNs) are introduced by~\cite{gori2005new} to study the structured data. Gori~\etal~\cite{gori2005new} extend GNNs from recursive neural networks (RNNs) to process graphs without losing topological information. 
There are two kinds of graph constructions~\cite{bruna2013spectral}, one based upon a hierarchical clustering of the domain, and the other based on the spectrum of the graph Laplacian.
Kipf~\etal~\cite{kipf2016semi} propose to encode the graph structure and features of nodes directly by a Graph Convolutional Network (GCN). Gilmer~\etal~\cite{gilmer2017neural} reformulate several existing graph models as a single common framework: Message Passing Neural Networks (MPNNs). Other graph models, such as  Graph Attention Network(GAT)~\cite{velivckovic2017graph}, EdgeConv~\cite{wang2019dynamic} and GraphSAGE~\cite{hamilton2017inductive}, are proposed to tackle the graph tasks.
In recent years, GNNs are also applied to many computer vision fields, including face clustering~\cite{wang2019linkage}, 3D person re-identification~\cite{zheng2020person} and point cloud classification~\cite{wang2019dynamic}. Some researchers have also applied graph models to the image retrieval tasks. For instance, similarity-Guided graph neural network (SGGNN)~\cite{shen2018person} is proposed to obtain similarity estimations and refine feature representations by incorporating graph computation in both training and testing stages. The group shuffling random walk (GSRW) layer ~\cite{shen2018deep} is integrated into deep neural networks for fully utilizing the affinity information between gallery images. Spectral feature transformation (SFT)~\cite{luo2019spectral}  incorporates spectral clustering technique into existing convolutional neural network (CNN) pipeline. To exploit the training data into the learning process, these graph-based methods integrate GNN into both the training and testing processes to utilize the relations between nodes. However, their performance is not as good as traditional re-ranking algorithms. The major limitation lies in the graph is build on a small set of data points, which discards massive contextual information in the total dataset. Therefore, the messages propagated by GNN are restricted in the local region and unable to make use of global information. Different from existing methods, we propose a GNN-based re-ranking method building on the entire dataset, capturing the topology structure of data.

\section{Our Approach}
\subsection{Overview} 

\noindent  \textbf{Problem Definition.} \label{sec:definition}
Given a query image $q$ and a gallery set with $n_g$
images $\mathcal{G} = \{g_i|i=1, 2,...,n_g\}$,
content-based image retrieval is to find relevant images among a large number of candidate images. Generally, we map images to a semantic feature space and sort gallery images according to the feature similarity with the query feature. The re-ranking approach is to further refine the initial retrieval results.

\subsection{Conventional Re-Ranking} \label{sec:rerank}
Re-ranking methods usually depend on additional criteria, such as neighbor similarity, to leverage the extra information between images. 
For example,~\cite{zhong2017re} employ $k$-reciprocal nearest neighbors to re-rank samples, which pull the relevant samples closer to each other. 
In this section, we briefly review this typical neighbor-based re-ranking method, which consists of two main steps. In the first step, it encodes the weighted $k$-reciprocal neighbor set into a $k$-reciprocal feature. In the second step, the $k$-reciprocal feature is improved by the local query expansion, fusing feature representation of neighbor samples. The final distance is calculated as the weighted sum of the original distance and the Jaccard distance.

\noindent  \textbf{Construction of $k$-reciprocal Neighbors.}
We define $\mathcal{N}(q, k)$ as the $k$-nearest neighbors (top-$k$ samples in the initial ranking list) of the query image $q$. $\mathcal{R}(q, k)$ is denoted as the $k$-reciprocal nearest neighbors of $q$ in the gallery, which can be formulated as:
\begin{equation}
\begin{split}
 \mathcal{R}(q, k) = \{g_i|(g_i \in \mathcal{N}(q, k))\cap (q \in \mathcal{N}(g_i, k))\}.
\end{split} 
\end{equation}
$\mathcal{R}(q, k)$ explicitly considers the neighbor of the nearest samples, enabling the cross-check of neighbor relations. Therefore, the $k$-reciprocal nearest neighbors can effectively reduce the noisy false-matches in the high-confidence candidates.
\textbf{To avoid the ambiguity, here we denote the number of the nearest neighbor for $k$-reciprocal calculation as $k_1$.}
Considering the severe visual appearance changes due to the pose and the occlusion,~\cite{zhong2017re} further introduce a refined expansion set $\mathcal{R^*}(q, k_1)$ by adding the $\frac{1}{2}k_1$-reciprocal
nearest neighbors of each candidate in $\mathcal{R}(q, k_1)$ as:
\begin{equation}
\label{eq:expand_k}
\begin{split}
& \mathcal{R}^*(q, k_1) \leftarrow \mathcal{R}(q, k_1) \cup  \mathcal{R}(g, \frac{1}{2}k_1) \\
& s.t. |\mathcal{R}(q, k_1) \cap  \mathcal{R}(g, \frac{1}{2}k_1)| 
\geq \frac{2}{3}|\mathcal{R}(g, \frac{1}{2}k_1)| \\ 
& \forall g \in \mathcal{R}(q, k_1),
\end{split}
\end{equation}
where $|\cdot|$ denotes the number of candidates in the set. Given the refined nearest neighbor set $\mathcal{R^*}(q, k_1)$, the neighbor information can be encoded as one $k$-reciprocal feature vector $F_{q} = [F_{q, g_1}, F_{q, g_2}, ..., F_{q, g_{n_g}}]$, where:

\begin{equation}
    F_{q, g_i}=\left\{
\begin{aligned}%
& exp({-d(q, g_i)}) & \text{if } g_i \in \mathcal{R}^*(q, k_1) \\
& 0 & \text{otherwise}
\end{aligned}
\right.,
\label{eq:k_reciprocal}
\end{equation}
and $d(q, g_i)$ represents the Mahalanobis distance~\cite{de2000mahalanobis} between query image $q$ and gallery image $g_i$. Different neighbors are treated distinctively based on neighbor relations and original pairwise similarities.

\noindent  \textbf{Local Query Expansion.}
Now we obtain the $k$-reciprocal feature vector for every image. We could apply the local query expansion~\cite{chum2007total} to further aggregate the similar features within $\mathcal{N}(q, k_2)$. It is worth noting that $k_2$ is different from $k_1$. $k_2$ represents the number of the nearest $k$-reciprocal feature neighbors for expansion. The final feature after the local query expansion can be formulated as:
\begin{equation}
    F_q = \frac{1}{k_2}\sum_{g_i \in \mathcal{N}(q, k_2)} F_{g_i}.
    \label{eq:query_expansion}
\end{equation}
We note that each neighbor is treated equally during aggregating process. After the expansion of $k$-reciprocal feature, the general Jaccard distance is defined as: 
\begin{equation}
    d_J(q, g_i) = 1 - \frac{\sum_{j=1}^{n_{g_n}} min(F_{q, g_j}, F_{g_i, g_j})}{\sum_{j=1}^{n_{g_n}} max(F_{q, g_j}, F_{g_i, g_j})}.
\end{equation}

Finally, distance $d^*$ combines the original distance and Jaccard distance to revise the initial ranking list:
\begin{equation}
    d^*(q, g_i) = (1 - \lambda)d_J(q, g_i) + \lambda d(q, g_i),
\end{equation}
where $\lambda \in [0, 1]$ denotes the penalty factor. The refined final distance is subsequently used to acquire the re-ranking list.
The $k$-reciprocal re-ranking significantly improves the mean average precision (mAP). It becomes a common post-processing component in image retrieval tasks. However, the $k$-reciprocal re-ranking is time-consuming since the massive set comparison operations in Eq.~\ref{eq:expand_k} are required to expand $k$-reciprocal neighbors.

\subsection{GNN-based Re-Ranking} \label{sec:gnn} 
\begin{figure}[t]
\begin{center}
   \includegraphics[width=1\linewidth]{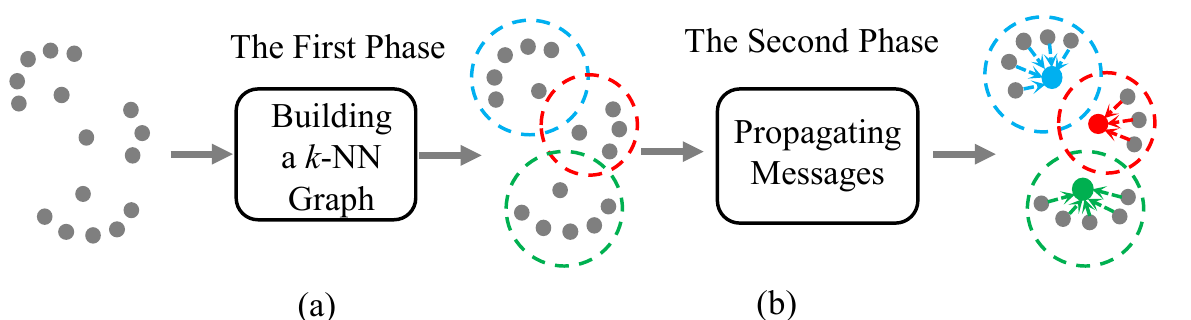}
\end{center}
   \caption{\textbf{The pipeline of our approach}.
   In the first phase~(left),  we build the $k$-NN graph, capturing the topology structure among data. For the second phase~(right), we employ the GNN to aggregate features from high-confidence samples~(nodes inside the dotted circle). The colored nodes represent the aggregated features.
   }
\label{fig:pipeline}
\end{figure}

In this section, we introduce the GNN-based re-ranking, which reduces the large time cost of complicated operations in conventional re-ranking methods. 
The key idea is that the similarity between images can be represented as a relation graph. 
Set comparison operations can be replaced by building a simple but discriminative graph, while features can be updated by the message propagation in GNN.
The proposed approach consists of the following two stages.
In the first stage, based on the entire image group (query images and gallery images), we construct a graph and encode local information in edges. In the second stage, the proposed GNN propagates messages by aggregating neighbor features with edge weights. The re-ranked retrieval list can be calculated by comparing the similarity of refined node features. The pipeline of our approach is shown in Figure~\ref{fig:pipeline}.

\noindent  \textbf{Construction of Graph.}
Formally, we denote $X_q$, $X_g$ and $X_u$ as original features of query set, gallery set and the union of query and gallery set, respectively. There are $n$ images in query and gallery set in total. 
Let $G = (\mathcal{V}, \mathcal{E})$ denote the graph where $\mathcal{V} = \{ v_1,..., v_{n}\}$ are the vertices and $\mathcal{E} \subseteq   \mathcal{V} \times \mathcal{V} $ are the edges. Each image is a vertex on the graph and connected edges represent the similarity between vertices.

First of all, cosine similarity matrix $\boldsymbol{S}$ is calculated:
\begin{equation}
\begin{split}
& \boldsymbol{S}_{ij} = cos(x_i, x_j),\\
& \text{where } x_i , x_j \in X_u, X_u = X_q \cup X_g.
\end{split} 
\end{equation}
Secondly, the features of neighbor relations are calculated.  
In $k$-reciprocal re-ranking,~\cite{zhong2017re} encode $k$-reciprocal feature by selecting candidates from a expansion of $k$-reciprocal neighbors. Masses of set intersection and comparison operations are required when selecting candidates, leading to huge time cost. 

We overcome this drawback by adopting a simple but effective strategy to obtain the contextual information of the whole image group. Corresponding to the $k$-reciprocal feature in re-ranking, we aim to encode node features by extracting neighbor information from $\boldsymbol{S}$ on the entire graph. Specifically, we first define the adjacent matrix $\boldsymbol{A}$ as
\begin{equation}
\boldsymbol{A_{i, j}}=\left\{
\begin{aligned}%
& 1 & \text{if } j \in  \mathcal{N}(i, k_1) \\
& 0 & \text{otherwise}
\end{aligned}
\right.,
\label{eq:a}
\end{equation}
where $\mathcal{N}(i, k_1)$  is the $k_1$ most similar candidates according to similarity matrix $\boldsymbol{S}$. 
Generally, an ideal adjacent matrix should be symmetric.
Further, the symmetric adjacent matrix $\boldsymbol{A}^*$ is introduced as:
\begin{equation}
\boldsymbol{A^*} = \frac{\boldsymbol{A} + \boldsymbol{A^T}}{2}.
\label{eq:a*}
\end{equation}
The value of $\boldsymbol{A}_{ij}^*$ is derived as:
\begin{equation}
\boldsymbol{A^*_{i, j}}=\left\{
\begin{aligned}%
& 1 & \text{if } j \in \mathcal{N}(i, k_1) \land i \in  \mathcal{N}(j, k_1)\\
& 0 & \text{if } j \notin  \mathcal{N}(i, k_1) \land i \notin  \mathcal{N}(j, k_1)\\
& 0.5 & \text{otherwise}
\end{aligned}
\right..
\end{equation}
We note that $\boldsymbol{A^*}$ encodes more adjacent information. It is because more neighbors are included and adaptive weights are assigned to high-confidence candidates. The experimental results in ablation study also show $\boldsymbol{A^*}$ performs better than $\boldsymbol{A}$. More importantly, due to the \textbf{simplicity}, \textbf{symmetry} and \textbf{sparsity} of $\boldsymbol{A^*}$, the calculation process can be highly parallelizable and efficient.

Here we define $h_i$ as the feature of vertex $v_i$, which can be extracted from the $i$-th row of the symmetric adjacent matrix $\boldsymbol{A^*}$:
\begin{equation}
    h_i = [\boldsymbol{A^*_{i,0}},...,\boldsymbol{A^*_{i,n}}].
    \label{eq:h_i}
\end{equation}
Such neighbor encoding feature performs better than the original feature because the hard negative images usually share one different neighbor set with the true-matches. 
Hence we construct node features based on neighbor similarity rather than directly adopting the original features. The comparison between different features can be found in the ablation study. 

Finally, we build a $k$-NN graph by connecting high-confidence edges to aggregate the feature of neighbors. Top-$k_2$ edges in $S$ for each vertex are connected and the  weights of edges are defined as:
\begin{equation}
e_{ij} = \boldsymbol{S}_{i,j}, j \in  \mathcal{N}(i, k_2).
\label{eq:e}
\end{equation}

\noindent  \textbf{Message Propagation.}
In the second phase, a feature aggregating process is required to further improve the retrieval performance, which is achieved by a local query expansion in conventional re-ranking methods. 
Similarly, in our GNN formulation, this process can be achieved by the message propagating approach~\cite{gilmer2017neural} on the graph. The key formula of this approach is described as below:
\begin{equation}
\label{eq:agg}
h^{(l+1)}_i = h^{(l)}_i + aggregate(\{f_{\Theta}(e_{ij}) \cdot h^{(l)}_{j}\}),
\end{equation}
where $h^{(l)}_i$ represents the feature of  $v_i$ in the $l$-th layer, $ f_{\Theta} $ is the function to compute the weight of propagating message and  $aggregate$ represents the aggregator types: $sum$, $mean$ or $max$.

We expect to find a suitable function $f_{\Theta}$, which can capture the relation between nodes by edge weights. With message propagation, high-confidence node features are enhanced and the unreliable node features are weakened.
Inspired by $\alpha$-weighted query expansion($\alpha$-QE)~\cite{radenovic2018fine}, we adopt $f_{\Theta}(e_{ij}) = e_{ij}^{\alpha}$, where $\alpha$ is a fixed value. 
Then the formula of our modified GNN can be refined as below:
\begin{equation}
\label{eq:h_agg}
h^{(l+1)}_i = h^{(l)}_i + aggregate(\{e_{ij}^{\alpha} \cdot h^{(l)}_{j},j \in \mathcal{N}(i, k_2) \}).
\end{equation}
  
Besides,  $h_{i}^{(l)}$ is regularized with $L_2$ norm after every message propagation on the graph. The last GNN layer will output the transformed node features $h^{(l)}_i$. Finally,
we derive the final ranking list according to the cosine similarity of refined features. 
Since the high-parallelism GNN propagates the message on the sparse graph efficiently, we can update all vertex features simultaneously.
The whole algorithm is summarized in Algorithm \ref{alg:Framwork}.

\begin{algorithm}[htb]  
  \caption{ Framework of our approach.}  
  \label{alg:Framwork}  
  \begin{algorithmic}[1]  
    \Require  
    The union of query and gallery features $X_u$; hyper-parameters $k_1$ and $k_2$  
    \Ensure  
      Final ranking list $L$
    \State Calculating the similarity matrix $\boldsymbol{S}$  according $X_u$
    \label{code:similarity}  
    \State Calculating adjacent matrix $\boldsymbol{A}$ using $k_1$ and $\boldsymbol{S}$
    \label{code:adj} 
    \State Calculating $\boldsymbol{A^*}$ accoding to Eq. \ref{eq:a*}
    \label{code:modify adj} 
    \State Deriving $h_i$ and $e_{ij}$ according to Eq. \ref{eq:h_i} and Eq. \ref{eq:e}
    \label{code:fram:add}  
    \State Building $k$-NN graph using $k_2$ 
    \label{code:fram:build graph}  
    \State Propagating message with GNN  
    \label{code:fram:propagate}  
    \State Calculating final ranking list $L$ according to the cosine similarity of refined node features  
    \label{code:fram:similarity} \\  
    \Return $L$;  
  \end{algorithmic}  
\end{algorithm}  

\subsection{Relation to Existing Methods}  
Our approach is related to two classes of approaches, re-ranking and GNN. The $k$-reciprocal re-ranking \cite{zhong2017re} can be viewed as a special case of our approach. We argue that the first phase equals to building the neighbor graph, while the second phase can be viewed as spreading message within the graph.  

In the first phase, the $k$-reciprocal feature is essential to calculate Jaccard distance in conventional re-ranking. However, the $k$-reciprocal feature requires to select the $k$-reciprocal neighbors of query as candidates, and then expand the candidates set by adding the common neighbors of query and candidates. 
The set comparison operations are not hardware acceleration friendly. It is due to the cross-check operation and a different number of $k$-reciprocal candidates during expanding the neighbor sets. 
In contrast, our method performs relation modeling by building a $k$-NN graph. It is natural to implement by matrix operations, which can be deployed on GPU easily. 
 
In the second phase, local query expansion equals to one layer GNN with $\alpha=0$ (in Eq.~\ref{eq:h_agg}) and the aggregator type sets to $mean$. 
The conventional query expansion message spreads with equal weights in a local region. In contrast, our approach spreads node features along with edge weights, combining both global and local information. We also enable different aggregate functions. In addition, as shown in experiments, the features of nodes can be further improved by aggregating features with the number of layers (\ie, two-layer GNN) increasing.

\section{Experiment}
We conduct experiments on five image retrieval datasets of different application scenarios, including a person re-identification dataset Market-1501~\cite{zheng2015scalable}, a vehicle re-identification dataset VeRi-776~\cite{liu2016deep}, two popular landmark retrieval datasets, \ie, Oxford-5k~\cite{philbin2007object} and Paris-6k~\cite{philbin2008lost}, and a drone-based geo-localization dataset University-1652~\cite{zheng2020university}. 
More details can be found in Section~\ref{sec:dataset}.
Besides, we provide an ablation study about the effect of different components, GNN layer number and hyper-parameters in Section~\ref{sec:ablation}. 
Finally, we discuss the  retrieval results on five datasets in Section~\ref{sec:retrieval performance}. 

\subsection{Datasets and Settings} ~\label{sec:dataset}

\noindent\textbf{Market-1501.} Market-1501~\cite{zheng2015scalable}
is collected in front of a supermarket in Tsinghua University by six cameras. It contains 32,668 images of 1,501 identities in total. Specifically, it consists of 12,936 training images with 751 identities
and 19,732 testing images with 750 identities. 3,368 images in the test set are selected as the query set. The images are detected by the DPM detector~\cite{felzenszwalb2009object}.

\noindent  \textbf{VeRi-776.} 
VeRi-776~\cite{liu2016deep} is collected 
in a real-world unconstrained traffic scene with various attribute annotations. Specifically, it contains 49,360 vehicle images of 776 vehicles captured by 20 cameras. 37,781 images of 576 vehicles are divided into the train set while 11,579 images of 200 vehicles are employed as the test set. The query set consists of 1,678 images.

\noindent  \textbf{Oxford-5k.} 
Oxford-5k~\cite{philbin2007object} is one of the widely-used landmark retrieval datasets. It contains 5,062 images collected from Flickr by searching particular Oxford landmark names. The collected images are manually annotated for 11 different landmarks. There are 55 query images, and the rest images are leverages as the gallery.

\noindent  \textbf{Paris-6k.} 
Similarly, Paris-6k~\cite{philbin2008lost} consists of 6,412 high resolution $(1024 \times 768)$ images collected from Flickr by searching particular Paris landmark names. There are 12 images in the query set.

\noindent  \textbf{University-1652.} 
University-1652~\cite{zheng2020university} contains 1,652 buildings of 72 universities around the world. There are 701 buildings of 33 universities in the training set, 701 buildings of the rest 39 universities in the test set, and 250 extra buildings serving as distractors in the gallery set. In particular, there are 37,855 drone-based images in the query set, and 951 satellite-based images in the gallery set.

\noindent \textbf{Evaluation Metrics.}
We mainly use mean average precision (mAP)~\cite{zheng2015scalable} to evaluate the performance. For each query, the average precision (AP) is calculated according to the area under the Precision-Recall curve. Then we calculate the mean value of APs of all queries, \ie,mAP, which takes both the precision and recall rate into consideration.
The Recall@K~\cite{zhai2017predicting, liu2019lending, vo2016localizing} denotes whether the top-$K$ images contain a true match. The value of Recall@K equals to 1 if the first matched image has appeared before the $K$-th image, which is sensitive to the position of the first matched image. 
On Market-1501 and VeRi-776, we also report the mean Recall@1 accuracy of all query images. For University-1652, according to the previous work~\cite{zhai2017predicting, liu2019lending, vo2016localizing}, we add the  Recall@1, Recall@5 and Recall@10 of the retrieval results. 

\noindent \textbf{Implementation Details.}
When testing, we use a two-layer GNN and set the aggregator as $sum$ function.
The parameter $\alpha$ in Eq.~\ref{eq:h_agg} is fixed as $2$.
For a fair comparison, we adopt the open-source baseline models to verify the effectiveness of the proposed method.
In particular, we employ a popular open-source person re-identification networks\footnote{\tiny\url{https://github.com/douzi0248/Re-ID}} as the Market-1501 baseline, which adopts a strong backbone, \ie, ResNet-50-ibn~\cite{pan2018two}, and fuses multi-branch information to enhance the feature representation. The final feature dimension is 1536. For VeRi-776, we deploy a vehicle re-identification model\footnote{\tiny\url{https://github.com/BravoLu/open-VehicleReID}}, which extracts the 2048-dimension feature.  
For Oxford-5k and Paris-6k, we similarly extract 2048-dimension visual feature from the open-source ResNet101-GeM\footnote{\tiny\url{https://github.com/filipradenovic/cnnimageretrieval-pytorch}}~\cite{radenovic2018fine} as the baseline.
For University-1652, we follow the official implementation of~\cite{zheng2020university}  as the baseline model\footnote{\tiny\url{https://github.com/layumi/University1652-Baseline}} and extract 512-dim visual feature. 
We note that all experiments are conducted on the same machine with one Intel(R) Xeon(R) CPU E5-2620 v2 @ 2.10Ghz, 164GB memory and one K40m GPU with 12GB memory.

\subsection{Ablation Studies} \label{sec:ablation}
\noindent\textbf{Effect of Different Components.} We study the mechanism of the proposed method on the Market-1501 dataset. 
We gradually add different components and choose different input node features to analyze the contribution of each part. As shown in the first two columns of Table~\ref{table:ablation}, we can see that the symmetric adjacent matrix $\boldsymbol{A^*}$ brings an improvement of 3.07\% in mAP  from 88.26\% to 91.33\%. 
With the help of message propagation, we achieve 2.87\% improvement in mAP and 1.97\% improvement in Recall@1 because the relevant samples are pulled much closer. Moreover, we observe that introducing edge weights improve the mAP from 94.20\% to 94.53\% and the Recall@1 from 95.64\% to 96.29\%, since nodes are treated differently according to the similarity.
Besides, using two-layer GNN further increases the mAP accuracy by 0.12\%. This suggests that aggregating more node features appropriately can improve the performance. 
Finally, we study different input features in the last three columns. It shows that node features derived from $\boldsymbol{A^*}$ outperform $\boldsymbol{A}$  and original features $x_i$. It is because the symmetric adjacent matrix feature contains the neighbor similarity instead of the original feature, which is more robust to the hard-negative.

\noindent\textbf{Effect of the GNN Layer Number.}\label{sec:layers} To take one step further, we study the impact of the different GNN layer numbers on the Market-1501 dataset. In Table~\ref{table_layer}, we report the Recall@1 and mAP with GNN from 1 layer to 256 layers.
With the number of layers increasing, the node feature gradually moves to the mean value of the local connected neighbors, compromising the ranking performance.
As a result, the Recall@1 reduces from 96.29\% to 93.91\%, and the mAP increases from 94.53\% to 93.21\% then reduces to 94.58\%. 
As shown in Table~\ref{table_layer}, the mAP converges to 93.21\%, when the number of the GNN layers is $256$. In this case, the proposed method is approximately equal to the average query expansion (AQE)~\cite{chum2007total}.
Because messages propagate again and again, yielding the over-smoothing result. In contrast, our method achieves the highest Recall@1 and mAP with one and two GNN layers. 

\begin{table}
\begin{center}
{
\small
\setlength{\tabcolsep}{1pt}
\begin{tabular}{l|ccccccc}
\toprule[1pt]
Components & \multicolumn{7}{c}{Performance} \\
\hline
Node features $(x_i)$ &$\checkmark$  & &  &  &$\checkmark$  & &\\
Node features $(\boldsymbol{A_i})$ & & & & & &$\checkmark$&\\
Node features $(\boldsymbol{A^*_i})$ &  &$\checkmark$ & $\checkmark$ & $\checkmark$ & & &$\checkmark$\\
Message propagation & & & $\checkmark$  & $\checkmark$ & $\checkmark$ & $\checkmark$&$\checkmark$\\
Edge weights  & & & & $\checkmark$  & $\checkmark$  & $\checkmark$&$\checkmark$\\ 
Two-layer GNN  & & & & &$\checkmark$& $\checkmark$ & $\checkmark$\\
\hline
mAP (\%) & 88.26 & 91.33 & 94.20 &  94.53 &93.83& 94.51 &  \textbf{94.65}\\
Recall@1 (\%) & 95.28 & 93.97 & 95.64 &\textbf{96.29} & 95.81 & 95.56 & 96.11 \\
\bottomrule[1pt]
\end{tabular}}
\end{center}
\caption{ \textbf{Ablation study.} We study different feature inputs, \ie, original feature $x_i$, adjacent matrix feature $\boldsymbol{A_i}$ and symmetric adjacent matrix feature $\boldsymbol{A^*_i}$. Message propagation means aggregating the feature of neighbor nodes by Eq~\ref{eq:agg}. Edge weights represent the weights of propagating message in Eq~\ref{eq:h_agg}. Two-layer GNN denotes the number of layers is 2.} 
\label{table:ablation}
\end{table}

\setlength{\tabcolsep}{15pt}
\begin{table}[t]
\small
\begin{center}
\begin{tabular}{l|cc}
\toprule[1pt]
\multirow{2}{*}{Methods}&  \multicolumn{2}{c}{Market-1501} \\ 
& Recall@1 (\%)      & mAP(\%)\\ \midrule[0.5pt]
\textbf{ours} (1 layer) &\textbf{96.29 }   & 94.53   \\
\textbf{ours} (2 layers) &96.11   & \textbf{94.65} \\
\textbf{ours} (4 layers) &95.49   & 94.58 \\
\textbf{ours} (8 layers) &95.96   & 94.45 \\
\textbf{ours} (16 layers) &95.21   & 94.18 \\
\textbf{ours} (32 layers) &94.92   & 93.90 \\
\textbf{ours} (64 layers) &95.01   & 93.67 \\
\textbf{ours} (128 layers) &94.44   & 93.33 \\
\textbf{ours} (256 layers) &93.91   & 93.21 \\
AQE~\cite{chum2007total} &95.64  &93.33 \\
\bottomrule[1pt]
\end{tabular}
\end{center}
\caption{\textbf{Effect of the GNN Layer Number.} Comparison with different number of GNN layers on the Market-1501 dataset. With the number of layers increasing, the mAP accuracy converges to 93.21\%. In this case, the proposed method is approximately equal to the average query expansion (AQE)~\cite{chum2007total}.}
\label{table_layer}
\end{table}

\begin{figure}[t]
\begin{center}
   \includegraphics[width=1\linewidth]{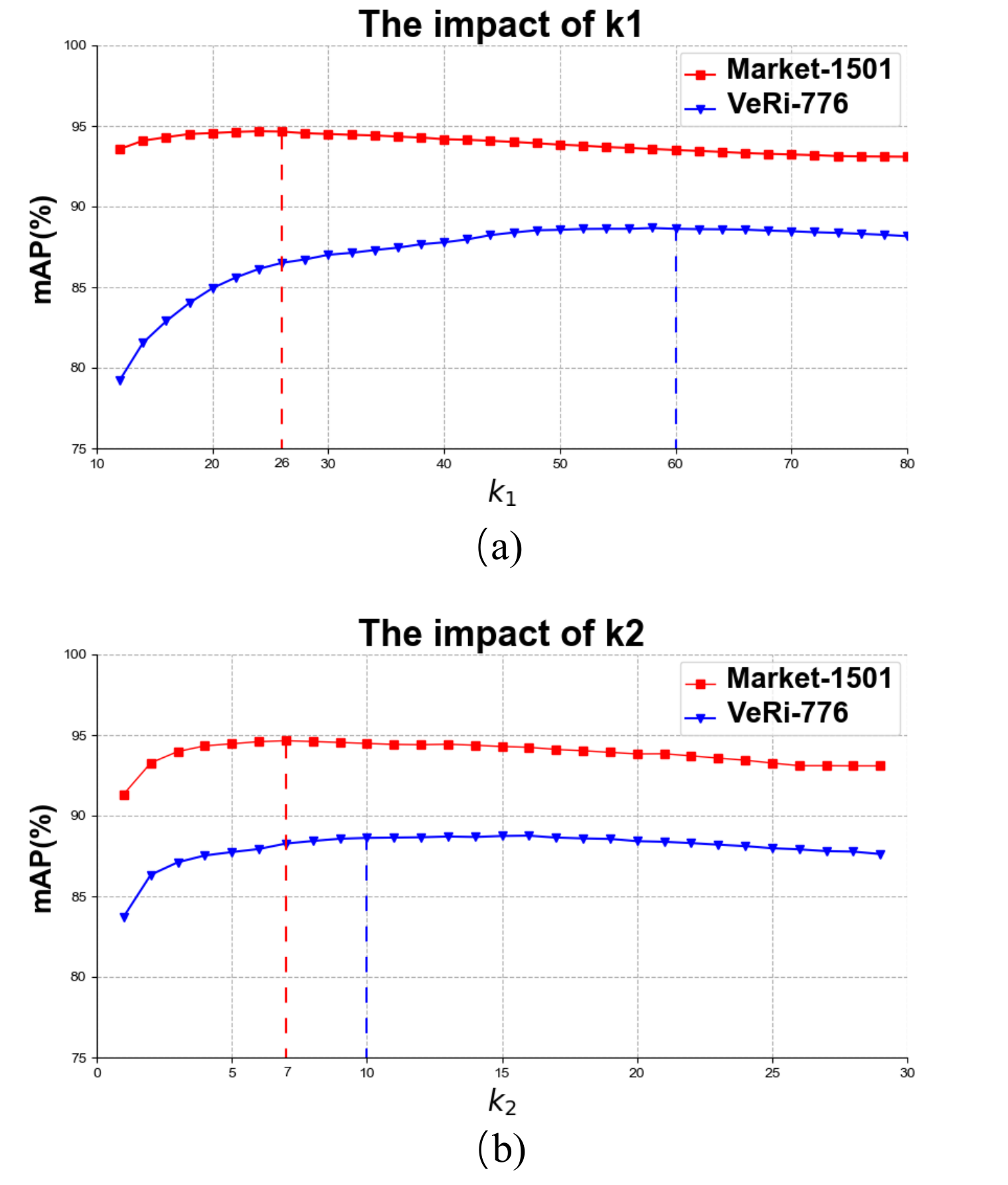}
\end{center}
  \caption{\textbf{The hyper-parameter analysis on $k_1$ and $k_2$}. We analyze hyper-parameters on Market-1501 and VeRi-776. For Market-1501, we, following the previous work ~\cite{van2020scan}, empirically fix $k_2 = 7$ in \textbf{(a)} to study $k_1$, and fix $k_1=26$ in \textbf{(b)} to study $k_2$. Similarly, for VeRi-776, we fix $k_2=10$ in \textbf{(a)} and $k_1=60$ in \textbf{(b)} to study the impact of hyper-parameters.
  }
\label{fig:k}
\end{figure}

\noindent\textbf{Effect of Hyper-parameter.}  \label{sec:hyper}
To analyze the impact of two hyper-parameters ($k_1$ and $k_2$), we conduct experiments on Market-1501 and VeRi-776 datasets. The hyper-parameter $k_1$ is used to calculate the adjacent matrix $\boldsymbol{A^*}$, while $k_2$ is the number of neighbors in message passing. 
Increasing the value of $k_1$ moderately introduces more neighbors.
$k_2$ is much smaller than $k_1$ because a large value may bring noisy nodes. 
According to the previous work~\cite{van2020scan}, the average number of images per class can be empirically estimated by $[\frac{n}{C}]$ ($C$ represents the number of classes, $[\cdot]$ indicates round down operation) and it is reasonable to construct neighbor relations based on this value, thus we set:
\begin{equation}
k_1= [\frac{n}{C}]. \label{eq:k_1}
\end{equation}
$k_2$ is estimated by a much smaller number than $k_1$ because a large value may bring noisy nodes. This selection rule is used in all following experiments. We first study the effect of $k_1$ by fixing the value of $k_2$. 
$k_2$ is set to 7 and 10 on Market-1501 and VeRi-776 respectively.
As shown in Figure~\ref{fig:k} (a), our approach is insensitive to $k_1$ in a wide range from 20 to 40 on Market-1501. For VeRi-776, the proposed method maintains a relatively high performance when $k_1$ changes from 50 to 70.
To evaluate the influence of  parameter $k_2$, we fix $k_1$ to 26 and 60 on Market-1501 and VeRi-776 separately.
Figure~\ref{fig:k} (b) shows the proformance of our approach is stable on Market-1501 when $k_2$ changes from $5$ to $10$.
Similar results can be achieved on VeRi-776 when keeping $k_1$ as 60 on VeRi-776 and changing $k_2$ from 10 to 15.
In general, the proposed method can achieve comparable results and is relatively robust to a large range of $k_1$ and $k_2$.

\setlength{\tabcolsep}{3pt}
\begin{table}[]
\scriptsize
\begin{center}
\begin{tabular}{l|c|ccc|cc}
\toprule[1pt]
\multirow{2}{*}{Methods}& \multirow{2}{*}{Platform} & \multicolumn{3}{c|}{Time} & \multicolumn{2}{c}{Performance}\\ 
&  & Phase 1    & Phase 2  &total & Recall@1 (\%) & mAP(\%)\\ \midrule[0.5pt]
$k$-reciprocal~\cite{zhong2017re} & CPU    &49.0s &40.2s   &89.2s  &94.65&96.11\\
\textbf{ours}(1 layer) & CPU   &1.2s   &29.4s  &30.6s & \textbf{96.29} & 94.53\\
\textbf{ours}(2 layers) & CPU   &1.2s   &58.7s  &59.9s&96.11&\textbf{94.65} \\
\textbf{ours}(2 layers) & GPU   &3.8ms   &5.6ms  &9.4ms &96.11&\textbf{94.65}\\
\bottomrule[1pt]
\end{tabular}
\end{center}
\caption{\textbf{Two-phase running time.} We provide the running time of each phase and compare the proposed GNN-based method with the k-reciprocal method. The one-layer GNN is used to make a comparison with $k$-reciprocal re-ranking, and we finally employ the two-layer GNN to achieve better performance.
We observe that our method performs better and is faster than $k$-reciprocal re-ranking on CPU.  We also report the time cost of our method operating on GPU with a high parallelism.  } 
\label{table:ablation_time}
\end{table}

\setlength{\tabcolsep}{14pt}
\begin{table*}[]
\begin{center}
\scriptsize
\begin{tabular}{l|c|c|c|c|c|c|c}
\toprule[1pt]
Methods & Platform& Market-1501 & VeRi-776 &Oxford-5k & Paris-6k & University-1652 & Average Time
 \\ \midrule[0.5pt]
AQE~\cite{chum2007total} & GPU & 3.4ms & 3.5ms   & 1.8ms & 1.7ms & 8.3ms & 3.7ms\\
$\alpha$-QE~\cite{radenovic2018fine} & GPU & 3.6ms & 3.7ms  & 1.9ms &1.6ms & 9.1ms & 4.0ms\\
\textbf{ours} & GPU  & 9.4ms & 5.2ms    &3.2ms & 3.5ms & 10.2ms & 6.3ms\\
\hline
SCA~\cite{bai2016sparse} & CPU & 43.2s & 18.4s   & 25.6s & 25.6s &89.7s  &40.5s \\
$k$-reciprocal~\cite{zhong2017re} & CPU & 89.2s &  36.7s  &64.1s& 64.4s & 135.9s & 78.1s
\\
\textbf{ours} & CPU & 60.0s & 21.3s & 17.1s& 23.2s & 85.5s & 41.4s \\
\bottomrule[1pt]
\end{tabular}
\end{center}
\caption{\textbf{Time cost.} We compare the proposed method with various post-processing methods on Market-1501, VeRi-776, Oxford-5k, Paris-6k and University-1652. We observe that our method finds one balance between speed and performance. For time cost on CPU, it is worth noting that proposed re-ranking method costs
the same time as SCA. Besides, our method achieves a similar speed with alpha-QE~\cite{radenovic2018fine} and AQE~\cite{chum2007total} but with large performance improvement on GPU.
}
\label{table:time}
\end{table*}

\setlength{\tabcolsep}{5pt}
\begin{table*}[]
\begin{center}
\scriptsize
\begin{tabular}{l|cc|cc|c|c|cccc}
\toprule[1pt]
\multirow{2}{*}{Methods}& 
\multicolumn{2}{c|}{Market-1501}&
\multicolumn{2}{c|}{VeRi-776} &
Oxford-5k&
Paris-6k& 
\multicolumn{4}{c}{University-1652}\\ 
& mAP (\%)  & Recall@1 (\%)   & mAP (\%)  & Recall@1 (\%)  & mAP (\%)& mAP (\%)&
mAP(\%) &Recall@1 (\%)    & Recall@5 (\%)     &Recall@10(\%)
\\ \midrule[0.5pt]
baseline&88.26  &95.28 & 78.94& 95.59& 88.21 & 92.62 &63.13&58.49&78.67&85.23     \\
AQE~\cite{chen2018group}&93.33   &95.64 & 82.49 & 89.22& 90.63 & 96.04 & 71.23& 67.62& 83.32&86.36\\
$\alpha$-QE~\cite{radenovic2018fine}   &93.51  &96.08& 82.77  & 89.72& 91.07 & 95.45& 71.69&68.18&83.66&86.71
\\
SCA~\cite{bai2016sparse}    &94.14  &96.08& 87.48 & \textbf{96.54} & 92.58 & 95.45 &74.11 & 70.52 & 86.22 & 90.34\\
$k$-reciprocal~\cite{zhong2017re}   &94.38  &95.90& 88.44 & 96.36& 92.31 & \textbf{96.49} & 73.67  &\textbf{70.71} & 83.86 & 85.65\\
\textbf{ours}& \textbf{94.65} & \textbf{96.11} & \textbf{88.61}& 96.42& \textbf{92.95}& 96.21 & \textbf{74.11} & 70.30 & \textbf{87.53} & \textbf{91.21}\\
\bottomrule[1pt]
\end{tabular}
\end{center}
\caption{\textbf{Retrieval performance.} We compare the proposed method with various post-processing methods on Market-1501, VeRi-776, Oxford-5k, Paris-6k and University-1652. mAP (\%) means average precision. We observe that the proposed method achieves the best or second-best performance on most datasets.}
\label{table:performance}
\end{table*}

\noindent\textbf{Time Cost Comparison.}  \label{sec:time cost}
We evaluate the efficiency of our method. We analyze the time cost of two phases on the Market-1501 dataset and compare our method with the conventional $k$-reciprocal re-ranking. Here we adopt the official implementation\footnote{\tiny\url{https://github.com/zhunzhong07/person-re-ranking}}.
As shown in Table~\ref{table:ablation_time}, we test the running time and performance of two methods on CPU. 
We can see that both the one-layer and two-layer GNN are faster than  $k$-reciprocal re-ranking in terms of the total time cost.
To further reduce the time cost, we extend the GPU-version of our method, which can accelerate the re-ranking process to \textbf{9.4ms}. To the best of our knowledge, there is no available GPU implementation of $k$-reciprocal re-ranking due to the sequential operations. Therefore, we do not include the GPU comparison with ~\cite{zhong2017re} but other methods \cite{chum2007total} and \cite{radenovic2018fine}. As shown in Table~\ref{table:time}  and  Table~\ref{table:performance}, the proposed method finds one balance point between speed and performance. 
In general, our method achieves better performance on most datasets. 
In terms of the time cost on CPU, it is worth noting that proposed re-ranking method achieves the similar speed with SCA~\cite{bai2016sparse}.
As for the GPU version, our method achieves 
a competitive speed with the vanilla methods including alpha-QE~\cite{radenovic2018fine} and AQE~\cite{chum2007total}, while we yield a better performance improvement.

\subsection{Retrieval Performance} \label{sec:retrieval performance}
\noindent\textbf{Experiments on Market-1501 and VeRi-776.}
As shown in Table~\ref{table:performance}, we compare the proposed method with other post-processing approaches on the two re-identification dataset, \ie, Market-1501 and VeRi-776.
There are two main observations. On the one hand, our method can improve mAP and Recall@1 by a large margin on the baseline. Specifically, our approach increases mAP by \textbf{6.39\%}  and Recall@1 by \textbf{0.83\%} on Market-1501.
On VeRi-776, we observe that the proposed approach gains 9.67\% improvement on mAP and 0.83\% improvement on the Recall@1.
On the other hand, we compare our approach with a variety of post-processing methods, including two feature similarity-based methods: AQE~\cite{chum2007total}, $\alpha$-QE~\cite{radenovic2018fine}, two neighbor similar-based methods:
SCA~\cite{bai2016sparse} and $k$-reciprocal~\cite{zhong2017re}. 
Results show that our approach outperforms all other methods both in mAP (\textbf{94.65\%}) and Recall@1 (\textbf{96.11\%}) on Market-1501. 
As for VeRi-776, our approach has achieved the highest mAP of \textbf{88.61\%} and second-best Recall@1 of 96.42\%. 
Besides, we also compare the time cost of different post-processing methods both on CPU and GPU platforms in Table~\ref{table:time}. 
The AQE~\cite{chum2007total} and $\alpha$-QE~\cite{radenovic2018fine} are implemented on GPU, while SCA~\cite{bai2016sparse} and $k$-reciprocal re-ranking~\cite{zhong2017re} operate on CPU due to the restriction of complex operations. The proposed method only takes about \textbf{9.4ms} and \textbf{5.2ms} to update  Market-1501 and VeRi-776 respectively, which is significantly better than the traditional neighbor-based re-ranking method, and competitive to the method based on feature similarity. 


\noindent\textbf{Experiments on Oxford-5k and Paris-6k.}
The proposed method is further evaluated on two small-scale landmark retrieval datasets, \ie, Oxford-5k and Paris-6k. 
The performances of different post-processing methods on ResNet101-GeM~\cite{radenovic2018fine} are reported in Table \ref{table:performance}. The feature dimension is 2048. Our approach has achieved the highest mAP accuracy of \textbf{92.95\%} on Oxford-5k, and competitive results 96.21\% mAP accuracy on Paris-6k. 
The experiment verifies the scalability of the method on small-scale datasets.  As shown in Table~\ref{table:time}, the re-ranking process only consumes \textbf{5.2ms}, which is faster than other re-ranking methods, \ie, SCA and k-reciprocal, and achieves a similar speed with the vanilla query expansion methods.

\noindent\textbf{Experiments on University-1652.}
We also provide experimental results on the drone-based geo-localization dataset~\cite{zheng2020university}. Given one drone-view image, the drone-view target localization task aims to find the corresponding satellite-view image to localize the target building in the satellite platform.
From Table \ref{table:performance}, we observe that our method can increase Recall@1 by 11.81\%, Recall@5 by 8.86\%, Recall@10 by 5.98\%, and mAP by 10.98\%. 
Comparing to other methods, our approach achieves the highest performance on Recall@5, Recall@10 and mAP with \textbf{87.53\%}, \textbf{91.21\%}, and \textbf{74.11\%}  respectively. From Table~\ref{table:time}, we can see that the proposed method only consumes \textbf{10.2ms} on GPU and 85.5s on CPU.

\section{Conclusion}
In this paper, we revisit re-ranking methods and identify the main challenge, \ie, high complexity. We re-formulate the re-ranking process as a graph neural network (GNN) function. Specifically, we explore the neighbor representation and deploy a two-layer GNN to aggregate the neighbor information of the entire data. The inherent attributes of GNN facilitate the parallelizable implementation and make it a hardware-friendly acceleration approach. 
Extensive experiments on five datasets demonstrate that our method is competitive to existing arts in terms of both retrieval precision and running time.
We hope this work can contribute to the future study of image retrieval tasks in real-world scenarios.




{\small
\bibliographystyle{ieee_fullname}
\bibliography{egbib}
}

\end{document}